\documentclass{article}
\usepackage{spconf,amsmath,graphicx}
\usepackage{amssymb,amsfonts}
\graphicspath{{figs/}}
\usepackage{booktabs}
\usepackage{etoolbox}
\usepackage{siunitx}
\usepackage{textcomp}
\usepackage{xcolor}
\usepackage{multirow}
\usepackage[caption=false, font=footnotesize]{subfig}
\usepackage[export]{adjustbox}
\usepackage{enumitem}
\usepackage{cite}

\title{TRANSFER LEARNING-BASED MODEL PROTECTION WITH SECRET KEY}

\name{MaungMaung AprilPyone and Hitoshi Kiya}
\address{Tokyo Metropolitan University, Tokyo, Japan}

\begin{document}
\maketitle

\begin{abstract}
We propose a novel method for protecting trained models with a secret key so that unauthorized users without the correct key cannot get the correct inference. By taking advantage of transfer learning, the proposed method enables us to train a large protected model like a model trained with ImageNet by using a small subset of a training dataset. It utilizes a learnable encryption step with a secret key to generate learnable transformed images. Models with pre-trained weights are fine-tuned by using such transformed images. In experiments with the ImageNet dataset, it is shown that the performance of a protected model was close to that of a non-protected model when the correct key was given, while the accuracy tremendously dropped when an incorrect key was used. The protected model was also demonstrated to be robust against key estimation attacks.
\end{abstract}

\begin{keywords}
Model Protection, Learnable Image Encryption, Model Watermarking
\end{keywords}

\section{Introduction}
\label{sec:intro}
Training successful deep neural networks (DNNs) is very expensive because it requires a huge amount of data and fast computing resources (e.g., GPU-accelerated computing). To train a convolutional neural network (CNN), for example, the ImageNet~\cite{ILSVRC15} dataset contains about 1.2 million images, and training on such a dataset takes days and weeks even on GPU-accelerated machines. In fact, collecting images and labeling them will also consume a massive amount of resources. Moreover, algorithms used in training a CNN model may be patented or have restricted licenses. Considering the expenses necessary for the expertise, money, and time taken to train a CNN model, a model should be regarded as a kind of intellectual property. While distributing a trained model, an illegal party may also obtain a model and use it for its own service.

To protect the copyrights of trained models, researchers have adopted digital watermarking technology to embed watermarks into the models~\cite{2017-ICMR-Uchida, 2020-NCA-Le, 2019-NIPS-Fan, 2019-MIPR-Sakazawa, 2018-Arxiv-Rouhani, 2018-ACCCS-Zhang, 2018-Arxiv-Chen, 2018-USENIX-Yossi}. These works focus on identifying the ownership of a model in question. However, a stolen model can be directly used by an attacker without arousing suspicion. Moreover, a stolen model can be exploited through model inversion attacks~\cite{2015-CCCS-Fredrikson} and adversarial attacks~\cite{2014-ICLR-Szegedy}. Therefore, a trained model should be protected against unauthorized access beyond ownership verification.

Recently, Fan et\ al.~\cite{2019-NIPS-Fan} proposed a passport-protected model-protection method. However, in their work, the network has to be modified with passport layers that use passports, and there are significant overhead costs in both training and inference time. Another work~\cite{pyone2020training} introduced a model protection method by taking inspiration from adversarial defenses~\cite{2020-ICIP-Maung,2020-Arxiv-Maung} that exploit the uniqueness of a key. They utilized a block-wise transformation with a key for model protection~\cite{pyone2020training}. However, it was tested only on CIFAR-10~\cite{2009-Report-Krizhevsky}, and the protected model was trained from scratch. Considering a large dataset like ImageNet~\cite{ILSVRC15}, it is not feasible to train a protected model from scratch as in~\cite{pyone2020training, 2019-NIPS-Fan}. Although, both the passport-protected~\cite{2019-NIPS-Fan} and key-protected~\cite{pyone2020training} methods train protected models from scratch, they do not consider transfer learning.

Transfer learning has been proved to be effective in various visual recognition tasks~\cite{kornblith2019better}. Transfer learning can be used in either of two scenarios: a pre-trained model can be transferred to a new model with the same number of classes or to a new one with a different number of classes (usually a lower number of classes). In this paper, we focus on the first scenario (i.e., transfer to the same number of classes) to confirm the effectiveness of the proposed method.

We propose a model protection method with a secret key that takes advantage of transfer learning for the first time. The proposed method also allows us to use a small subset of a training dataset to replace an unprotected model with a protected one. In addition, it does not need to modify a network, and therefore, there is no overhead for both training and inference time. In an experiment on ImageNet, the performance of a model protected by the proposed method is demonstrated not only to be close to that of a non-protected one when the key is correct but also to significantly drop when using an incorrect key.

\section{Related Work}
\label{sec:related-work}
\subsection{Model Watermarking}
There are mainly two categories of DNN model watermarking: white-box and black-box. A white-box approach requires access to model weights for embedding and extracting a watermark as in~\cite{2017-ICMR-Uchida,2018-Arxiv-Chen,2018-Arxiv-Rouhani,2019-NIPS-Fan}. In contrast, black-box approaches~\cite{2018-USENIX-Yossi,2018-ACCCS-Zhang,2019-NIPS-Fan,2019-MIPR-Sakazawa,2020-NCA-Le} do not need to access model weights and observe the input and output of a model in doubt to verify the ownership of the model.

These existing model-watermarking schemes focus on ownership verification only. Thus, a stolen model can be directly used and exploited without arousing suspicion because the performance of a protected model (i.e., fidelity) is independent of the embedded watermark.

In addition, Fan et\ al.~\cite{2019-NIPS-Fan} pointed out that conventional ownership verification schemes are vulnerable against ambiguity attacks~\cite{1998-IEEEJSAC-Craver} where two watermarks can be extracted from the same protected model, causing confusion regarding ownership. Therefore, Fan et\ al.~\cite{2019-NIPS-Fan} introduced passports and passport layers. The passports in~\cite{2019-NIPS-Fan} are a set of extracted features of a secret image/images or equivalent random patterns from a pre-trained model, and the passport layers are additional layers in the network. Therefore, there are significant overhead costs in both the training and inference phases, in addition to user-unfriendly management of lengthy passports in~\cite{2019-NIPS-Fan}. Moreover, the protection method with passports~\cite{2019-NIPS-Fan} was evaluated only on CIFAR datasets~\cite{2009-Report-Krizhevsky} and does not consider transfer learning.

\subsection{Learnable Image Encryption}
Learnable image encryption perceptually encrypts images while maintaining a network's ability to learn the encrypted ones for classification tasks. Most early methods of learnable image encryption were originally proposed to visually protect images for privacy-preserving DNNs~\cite{2018-ICCETW-Tanaka,2019-Access-Warit,2019-ICIP-Warit,sirichotedumrong2020gan,ito2020image,ito2020framework}.

Recently, adversarial defenses in~\cite{2020-ICIP-Maung,2020-Arxiv-Maung} also utilized learnable image encryption methods. Here, instead of protecting visual information, these works focus on controlling a model's decision boundary with a secret key so that adversarial attacks are not effective on such models trained by learnable transformed images.

Another use case of learnable image encryption is the model protection proposed in~\cite{pyone2020training}. However, the method in~\cite{pyone2020training} requires training a protected model from scratch and never considers transfer learning. In this paper, we adopt a block-wise image encryption method as in~\cite{2020-ICIP-Maung,2020-Arxiv-Maung} to transform images prior to training and testing.

\section{Proposed Method}
\subsection{Overview}
\begin{figure}[tbp]
\centerline{\includegraphics[width=\linewidth]{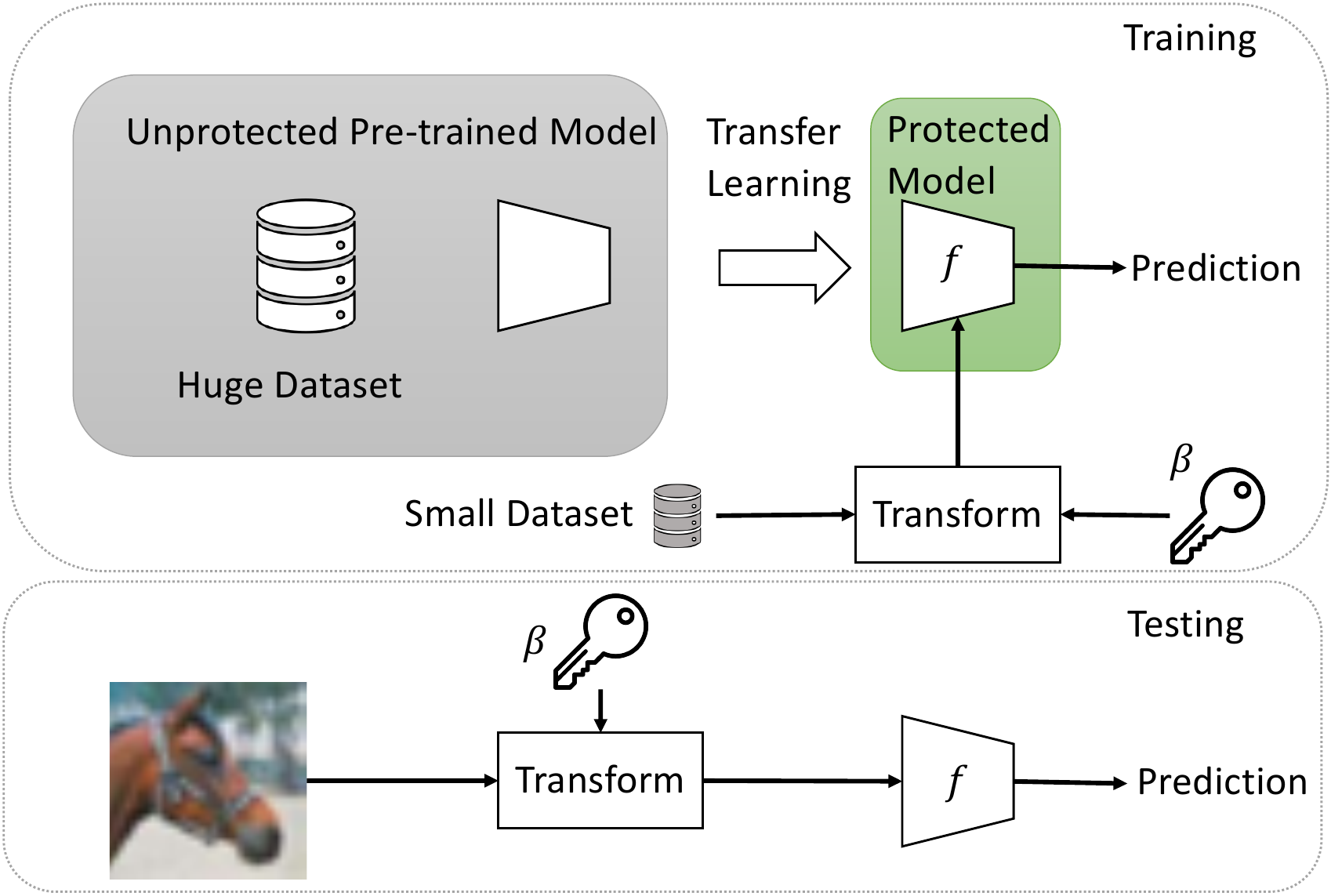}}
\caption{Image classification with proposed model-protection method\label{fig:overview}}
\end{figure}

An overview of image classification with the proposed method is depicted in Fig.~\ref{fig:overview}. In the proposed model protection, a model $f$ is not trained from random weights. Instead, $f$ is trained by taking advantage of transfer learning. In other words, $f$ is trained by fine-tuning pre-trained weights with a small dataset that consists of input images transformed by using secret key $\beta$. The resulting $f$ is protected by key $\beta$. For testing, test images are also transformed with the same key $\beta$ before testing. Therefore, the authorization of model $f$ is verified upon secret key $\beta$ during model inference.

\subsection{Block-wise Transformation}
\label{sec:transformation}
We use negative/positive transformation with a secret key to transform input images before training or testing a protected model as well as in~\cite{2020-Arxiv-Maung}. The following are steps for transforming input images, where $c$, $w$, and $h$ denote the number of channels, width, and height of an image tensor $x \in {[0, 1]}^{c \times w \times h}$.

\begin{figure}[!t]
\centering
\subfloat{\includegraphics[width=0.3\linewidth]{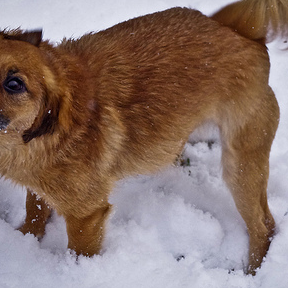}%
\label{fig:193}}
\hfil
\subfloat{\includegraphics[width=0.3\linewidth]{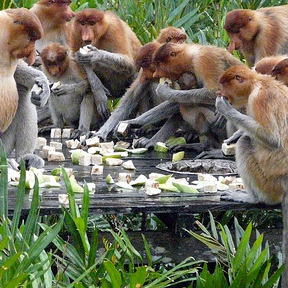}%
\label{fig:376}}
\hfil
\subfloat{\includegraphics[width=0.3\linewidth]{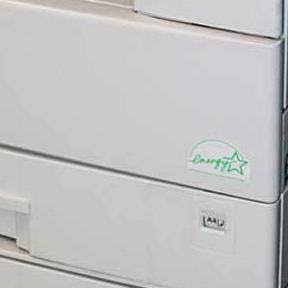}%
\label{fig:713}}\\
\renewcommand{\thesubfigure}{a} 
\subfloat[Australian Terrier]{\includegraphics[width=0.3\linewidth]{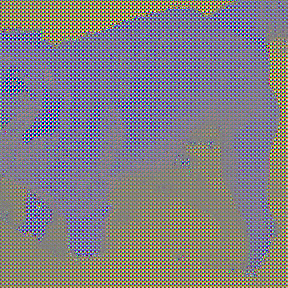}%
\label{fig:trans-193}}
\hfil
\renewcommand{\thesubfigure}{b}
\subfloat[Proboscis Monkey]{\includegraphics[width=0.3\linewidth]{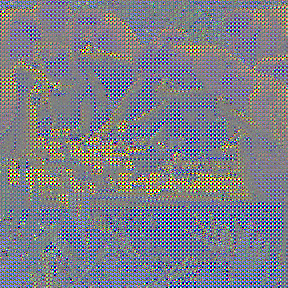}%
\label{fig:trans-376}}
\hfil
\renewcommand{\thesubfigure}{c}
\subfloat[Photocopier]{\includegraphics[width=0.3\linewidth]{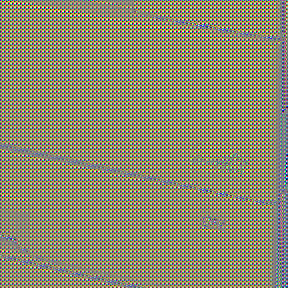}%
\label{fig:trans-713}}\\
\caption{Example of block-wise transformed images ($M = 4$) with key $\beta$ (second row). Images in first row are original.\label{fig:images}}
\end{figure}

\begin{enumerate}
\item Divide $x$ into blocks with a size of $M$ such that $\{B_{(1,1)}, \ldots, B_{(\frac{w}{M}, \frac{h}{M})}\}$.
\item Transform each block tensor $B_{(i, j)}$ into a vector $b_{(i,j)} = [b_{(i,j)}(1), \ldots, b_{(i,j)}(c \times M \times M)]$.
\item Generate a key $\beta$, which is a binary vector, i.e.,
 \begin{equation}
 \beta = [\beta_1, \dots, \beta_k, \dots, \beta_{(c\times M \times M)}], \beta_k \in \{0, 1\},
 \end{equation}
where the value of the occurrence probability $P(\beta_k)$ is $0.5$.
\item Multiply each pixel value in $b_{(i, j)}$ by $255$ to be at $255$ scale with 8 bits.
\item Apply negative/positive transformation to every vector $b_{(i, j)}$ with $\beta$ as
 \begin{equation}
 b'_{(i, j)}(k) = \left\{
 \begin{array}{ll}
 b_{(i, j)}(k) & (\beta_k = 0)\\
 b_{(i, j)}(k) \oplus (2^L - 1) & (\beta_k = 1),
 \end{array}
 \right.
 \end{equation}
where $\oplus$ is an exclusive or (XOR) operation, $L$ is the number of bits used in $b_{(i, j)}(k)$, and $L = 8$ is used in this paper.
\item Divide each pixel value in $b'_{(i, j)}$ by $255$ to be at $[0, 1]$ scale.

\item Integrate the transformed vectors to form an image tensor $x' \in {[0, 1]}^{c \times w \times h}$.
\end{enumerate}
An example of images (three different classes from the ImageNet test set) transformed by negative/positive transformation with $M = 4$ is shown in Fig.~\ref{fig:images}.

\subsection{Transfer Learning}
In practice, CNNs are not trained from the beginning with random weights because creating a large dataset like ImageNet is difficult and expensive. Therefore, CNNs are usually pre-trained with a larger dataset~\cite{2015-ICLR-Simonyan}. There are two major transfer-learning scenarios:
\begin{itemize}
\item \textbf{Fixed CNN:} A pre-trained CNN model is used as a fixed feature extractor, and the last fully connected layer is replaced with a targeted number of classes. In other words, convolutional layers are frozen, and only the last fully connected layer is trained with random initialization from scratch.
\item \textbf{Fine-tuned CNN:} In this scenario, the CNN is fine-tuned from a pre-trained model. Here, it is possible that some convolutional layers can be fixed or the whole CNN is fine-tuned.
\end{itemize}

In this paper, we fine-tuned a whole CNN with a small dataset comprised of learnable transformed images with a secret key in order to protect a model.

\subsection{Key Estimation Attack}
We consider a threat model where a model is stolen and transformation details are known except for the secret key. In this scenario, an attacker may try all possible keys (brute-force attack). The key space $\mathcal{K}$ of negative/positive transformation is given by
\begin{equation}
 \mathcal{K}(c \times M \times M) = 2^{(c \times M \times M)}.
\end{equation}
Therefore, the key space will vary with respect to block size $M$.

However, checking all possible keys may not be feasible, and the attacker may estimate a key heuristically by observing the accuracy of his/her test dataset. Elements in $\beta$ can be rearranged in accordance with the improvement in accuracy. We simulate this attacking scenario by swapping values in each index pair of $\beta$ if the accuracy improves.

Key estimation attacks do not guarantee that the attacker will find the correct key because the attacker does not know the actual performance of the correct key. The robustness of the proposed method against key estimation attacks will be demonstrated in the following section on experiments.

\section{Experiments}
\subsection{Setup}
We utilized the ImageNet dataset~\cite{ILSVRC15}, which comprises 1.28 million color images for training and 50,000 color images for validation. We progressively resized images during training, starting with larger batches of smaller images to smaller batches of larger images. We adapted three phases of training from the DAWNBench top submissions as mentioned in~\cite{2020-ICLR-Wong}. Phases 1 and 2 resized images to 160 and 352 pixels, respectively, and phase 3 used the entire image size from the training set. The augmentation methods used in the experiment were random resizing, cropping (sizes of $128$, $224$, and $288$, respectively, for each phase), and random horizontal flip. Both training and testing images were transformed with negative/positive transformation with a block size $M = 4$.

We deployed deep residual networks~\cite{2016-CVPR-He} with 50 layers (ResNet50) with pre-trained weights and fine-tuned for $15$ epochs with cyclic learning rates~\cite{2017-Arxiv-Smith} and mixed precision training~\cite{2017-Arxiv-Micikevicius}. We adapted the training settings from~\cite{2020-ICLR-Wong} with the removal of weight decay regularization from batch normalization layers.

\subsection{Classification Performance}
Table~\ref{tab:results} summarizes the classification results for protected models and a baseline (unprotected one). We fine-tuned models by using subsets of the training dataset with \SI{10}{\percent}, \SI{20}{\percent}, \SI{30}{\percent}, and \SI{100}{\percent} of the training set, respectively. Images in the sub-datasets were transformed by negative/positive transformation with a secret key $\beta$ and $M = 4$ as mentioned in Section~\ref{sec:transformation}. We tested the proposed method under three conditions: with a correct key $\beta$, with an incorrect key $\beta'$, and with plain images.

When a correct key $\beta$ was given, the model trained with a \SI{10}{\percent} dataset had about \SI{9}{\percent} less accuracy than that of the baseline. However, when the whole dataset was used, the accuracy was almost the same as the baseline accuracy (i.e., \SI{72.63}{\percent}). Even when the whole dataset was used, transfer learning significantly reduced the training time because the model was trained only for 15 epochs. When using an incorrect key $\beta'$ or plain images, the accuracy was extremely low, suggesting the strength of the proposed method against unauthorized access.

\robustify\bfseries
\sisetup{table-parse-only,detect-weight=true,detect-inline-weight=text,round-mode=places,round-precision=2}
\begin{table}[tbp]
 \caption{Accuracy (\SI{}{\percent}) of protected models and baseline model for ImageNet\label{tab:results}}
 \centering
 \begin{tabular}{lSSS}
 \toprule
 {Model} & {Correct ($\beta$)} & {Incorrect ($\beta'$)} & {Plain}\\
 \midrule
 {\SI{10}{\percent} dataset} & 64.134 & 0.16 & 0.3\\
 {\SI{20}{\percent} dataset} & 67.454 & 0.25 & 1.04\\
 {\SI{30}{\percent} dataset} & 68.866 & 0.24 & 0.73\\
 {\SI{100}{\percent} dataset} & \bfseries \num{72.63} & 0.69 & 0.36\\
 \midrule
 {Baseline} & \multicolumn{3}{c}{73.70 (Not protected)}\\
 \bottomrule
 \end{tabular}
\end{table}

\subsection{Robustness Against Key Estimation Attack}
We also evaluated the proposed method in terms of robustness against key estimation attacks. Table~\ref{tab:key} captures the results for all protected models under the use of a subset of the training dataset with various sizes. The model trained by the smallest dataset (i.e., \SI{10}{\percent}) had the lowest accuracy, and that trained by the whole dataset had a \SI{25.43}{\percent} accuracy. All in all, the key estimation attack did not guarantee that a good enough key would be found, and the performance accuracy was not usable, suggesting the robustness of the proposed method.

\robustify\bfseries
\sisetup{table-parse-only,detect-weight=true,detect-inline-weight=text,round-mode=places,round-precision=2}
\begin{table}[tbp]
 \caption{Accuracy (\SI{}{\percent}) of protected models against key estimation attack for ImageNet\label{tab:key}}
 \centering
 \begin{tabular}{lS}
 \toprule
 {Model} & {Estimated ($\beta'$)}\\
 \midrule
 {\SI{10}{\percent} dataset} & 0.17\\
 {\SI{20}{\percent} dataset} & 1.61\\
 {\SI{30}{\percent} dataset} & 9.42\\
 {\SI{100}{\percent} dataset} & 25.43\\
 \bottomrule
 \end{tabular}
\end{table}

\subsection{Functional Comparison with State-of-the-art Methods}
To the best of our knowledge, there are only two methods~\cite{pyone2020training, 2019-NIPS-Fan} where the protection method is directly dependent on model performance (i.e., key/passports protected models). However, both of them were not tested on ImageNet, and it is not feasible to train an ImageNet model from scratch. The other model watermarking methods such as~\cite{2017-ICMR-Uchida,2019-NIPS-Fan,2018-Arxiv-Chen, 2018-Arxiv-Rouhani,2018-USENIX-Yossi, 2018-ACCCS-Zhang,2019-MIPR-Sakazawa, 2020-NCA-Le} focus on ownership verification only when a stolen model is in question. Therefore, the embedded watermark is independent of model accuracy.

Since the state-of-the-art model-protection methods~\cite{pyone2020training, 2019-NIPS-Fan} cannot be directly compared with the proposed method as described above, we performed a functional comparison with a key-protected method (Pixel Shuffling)~\cite{pyone2020training} and a passport-protected method (Scheme $\mathcal{V}_1$)~\cite{2019-NIPS-Fan}. Both methods control the accuracy of performance by using a key or passports and have low performance degradation. However, scheme $\mathcal{V}_1$~\cite{2019-NIPS-Fan} has to modify a network with additional passport layers; therefore, it introduces overheads in training (\SI{15}{\percent}--\SI{30}{\percent}) and inference (\SI{10}{\percent}) processes as mentioned in~\cite{2019-NIPS-Fan}. In contrast, the proposed method does not introduce any overhead during training and testing, in addition to being applicable to transfer learning.

\robustify\bfseries
\sisetup{table-parse-only,detect-weight=true,detect-inline-weight=text,round-mode=places,round-precision=2}
\begin{table}[tbp]
 \caption{Functional comparison of proposed method and state-of-the-art methods\label{tab:comparison}}
 \centering
 \resizebox{\columnwidth}{!}{%
 \begin{tabular}{lcccc}
 \toprule
 \multirow{2}{*}{Model} & {Performance} & {Network} & {Performance} & \multirow{2}{*}{Overhead}\\
 & {Dependency} & {Modification} & {Degradation} & \\
 \midrule
 {Scheme $\mathcal{V}_1$~\cite{2019-NIPS-Fan}} & Passports & Yes & Low & Significant\\
 {Pixel Shuffling~\cite{pyone2020training}} & Key & No & Low & No\\
 {Proposed Method} & Key & No & Low & No\\
 \bottomrule
 \end{tabular}
}
\end{table}

\section{Conclusion}
In this paper, we proposed a model protection method in which a model is fine-tuned with a subset of a training dataset. Images in the sub-datasets are transformed by a block-wise transformation with a secret key prior to training and testing a model. The performance accuracy of a protected model trained by using \SI{10}{\percent} of a training dataset was about \SI{9}{\percent} less than that of a baseline model. When using the whole dataset, the accuracy was close to the baseline accuracy, and transfer learning significantly reduced the training time because the model was trained only for 15 epochs. The proposed model-protection method was also confirmed to be robust against key estimation attacks and not usable when using an incorrect key or plain images. Moreover, it does not introduce any overhead in both training and inference time.

\begin{small}
\bibliographystyle{IEEEbib}
\bibliography{refs}
\end{small}

\end{document}